# Paging Dr. GPT: Extracting Information from Clinical Notes to Enhance Patient Predictions

David Anderson, PhD
Management and Operations
Villanova School of Business

Michaela Anderson, MD
Division of Pulmonary, Allergy, and Critical Care Medicine
University of Pennsylvania

Margret V. Bjarnadottir, PhD
Decisions, Operations & Information Technology
University of Maryland,
7621 Mowatt Ln,
College Park, MD, 20742, US.

Stephen Mahar, PhD
Management and Operations
Villanova School of Business

Shriyan Reyya,
Decisions, Operations & Information Technology
University of Maryland

## Abstract

There is a long history of building predictive models in healthcare using tabular data from electronic medical records. However, these models fail to extract the information found in the unstructured clinical notes, which document diagnosis, treatment, progress, medications, and care plans—factors. In this study, we investigate how answers generated by GPT-4o-mini (ChatGPT) to simple clinical questions about patients, when giving access to the patient's discharge summary can support patient-level mortality prediction. Using data from 14,011 first-time admissions to the Coronary Care or Cardiovascular Intensive Care Units in the MIMIC-IV Note dataset, we implement a transparent framework that uses GPT responses as input features in logistic regression models. Our findings demonstrate that GPT-based models alone can outperform models trained on standard tabular data, and that combining both sources of information yields even greater predictive power, increasing AUC by an average of 5.1 percentage points and increasing PPV by 29.9% for the highest-risk decile. These results highlight the value of integrating large language models (LLMs) into clinical prediction tasks and underscore the broader potential for using LLMs in any domain where unstructured text data remains an underutilized resource.

# 1. Introduction

Healthcare systems are under mounting pressure due to growing patient volumes, constrained resources, and rising costs. In this difficult environment, the ability to effectively forecast patient outcomes is increasingly critical for appropriate resource allocation to improve patient care and healthcare delivery. It is therefore unsurprising that over the past few decades the academic literature has exploded with machine learning models for different disease and healthcare contexts[1-5]. A large part of this literature has focused on utilizing tabular data or on a single modality such as imaging. More recently, multi-modal machine learning frameworks[6] have been introduced in order to take advantage of the many available data modalities, images, lab outcomes, diagnoses and tests.

In this work we focus on information from electronic medical records (EMRs) for patient level predictions, which has a long tradition of using machine learning models to predict healthcare outcomes (e.g., [7-10]). While EMRs are known for providing valuable structured data, this data often fails to capture the full scope of the patient's condition. Meanwhile, much information about patients is seen only by their healthcare providers and resides in unstructured discharge summaries, where providers record nuanced observations and insights. In fact, unstructured data, which includes clinical notes and patient reports, makes up approximately 80% of the EMR[11,12]. Although such data are often qualitative and present interpretation challenges, natural language processing and machine learning have shown the potential for integrating such complex unstructured data into prediction pipelines for improved performance (e.g., in the context of patient safety[13]).

Traditionally, unstructured text data has been underutilized in predictive modeling due to the challenges involved in making insight from the text available to the model. As a result, key information that could improve predictions of patient outcomes often remains untapped. However, recent advancements in large language models (LLMs) are poised to have a transformational impact[14,15]. LLMs have already demonstrated significant capabilities and are set to revolutionize natural language processing. These advancements will change how we leverage textual data in healthcare. *In this paper we* *introduce an expert driven approach for incorporating LLMs into standard machine learning pipelines to improve predictions in the health care context.*

In contrast to traditional natural language processing approaches used in the past[16-20], *we use an LLM to answer clinician questions about the patient based on their discharge summary*. We then train standard prediction models based on those answers. The prompt (the questions posed to GPT) asks the LLM to "read" the patient's discharge note and respond to three questions summarizing the patient's medical condition: What is the patient's risk of death? What is the patient's risk of readmission? How would you rate the patient's overall health? Figure 1 summarizes our modeling approach.

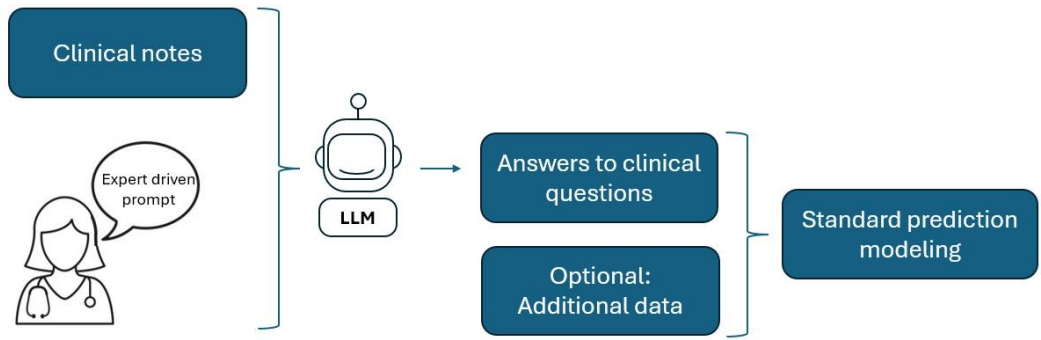

Figure 1. Embedding LLMs into a standard machine learning pipeline using an expert-guided LLM prompt. With access to a large set of clinical notes, LLM's answers to clinical questions about each patient are used in standard machine learning models.

**2. Methods**

Patient Population and outcomes

This study uses the MIMIC-IV Note dataset[21,22]. To construct a relatively homogeneous patient population, we include first admissions of patients with heart-specific medical issues, namely those admitted to the Coronary Care Unit (CCU) or Cardiovascular Intensive Care Unit (CVICU). We exclude patients who died before discharge, as the goal is to predict future mortality. The final cohort has 14,011 patients; each is either in the CCU or CVICU, is discharged alive, and has a discharge summary. We then study three outcomes, 90-day mortality, 1-year mortality and 90-day readmissions.

EMR Tabular Feature Extraction

Our EMR features are based on prior literature[23-26] and focus on patient demographics, laboratory tests, and comorbidity status. Features with multiple values per admission, such as vital signs and test results, are collapsed into minimum, maximum, and mean values over the inpatient stay. The full list of the tabular data features is included in eTable 1 in the supplement.

Discharge Summary Features

We interact with GPT through a prompt, a text or input to GPT asking it to generate a response. GPT reads the prompt and then responds based on what it understands from it. Prompt engineering has emerged as one of the key considerations in LLM adoption, as the structure and content of a prompt directly impacts the quality of the response.

We used ChatGPT 4o-mini via the OpenAI API, and tested our prompt through an iterative process of creating queries and testing them on a sample of 100 random clinical notes. Initially our prompt included specific questions about each organ system. In our numerical experiments, we however found that simply asking three questions about the patients' health, mortality and readmission risk resulted in equally good predictions. The finalized prompt for each discharge note has four key parts: i) the context, ii) the clinical note itself, iii) the three questions, and iv) instructions on how to format the answers (to ensure consistency in the output format, which is critical when featurizing data at scale). The full prompt is:

> *Please read the following physician discharge note and answer the questions listed below.*
> *The note text is: [the note text]*
>
> *Questions:*
> *1. What is the patient's risk of death? (Rate no risk = 1 to very high risk = 100)*
> *2. What is the patient's risk of readmission? (Rate no risk = 1 to very high risk = 100)*

*3. How would you rate the patient's overall health? (Rate very ill = 1 to perfect = 100)*

*Instructions:*
*Provide your answers as a semicolon-delimited list in the same order as the questions.*

*Example: 1. #; 2. #; 3. #;*

<u>Models and Model Evaluation</u>

The dataset is divided into a 70% training set and a 30% testing set, and all results are reported on the test set. To evaluate the usefulness of GPT's answers, we train logistic regression models using LASSO regularization for three outcomes: 90-day and 1-year mortality, and 90-day readmission. The models are implemented with the glmnet package in R[27], and tuned by selecting the shrinkage parameter that minimizes cross-validation error in the training set. For each outcome, we use three different feature sets: i) 70 EMR features (the full list of features is available in eTable 1 in the supplement), ii) GPT's answers to the three questions (GPT's estimation of risk of death, risk of readmission, and overall health), and iii) both the 70 EMR and the three GPT answers. We use the area under the ROC curve (AUC) as a performance indicator.

We further study two other performance measures, the overall error rate (1-accuracy) and the positive predictive value of each model and outcome. Specifically, we report the positive predictive value when the top 2.5%, 5%, 10% and 20% of the population are labeled as high risk. Finally, we both compare the risk scores (predicted probability of the outcomes) based on the different models to understand the differences in the patient predictions from different models and then examine one of the GPT variables in more detail to understand

The study protocol was reviewed by Villanova University's Institutional Review Board and given an exempt status. The instance of GPT used was hosted on an internal Villanova Microsoft Azure server and was opted out of human review of the conversation, following the responsible use guidelines: https://physionet.org/news/post/gpt-responsible-use.

**3. Results**

The results show that a simple model using only GPT's replies for a patient's risk of readmission, risk of death, and overall health scores consistently outperforms a model based on the 70 EMR-derived features. Further, the model using both EMR and GPT features outperforms the individual (i.e., EMR-only and GPT-only) models, adding an average of 3.7 percentage points to the AUC and reducing the overall classification error by an average of 22.2%. Table 1 summarizes the accuracy and AUC values for the EMR-only model, the GPT-only model, and the GPT+EMR model for all outcomes. We observe that the AUC of the readmission models are lower, which is in part driven by low prevalence (6%) of readmission in our population. We further note that the AUC of the models using both feature sets are, to the best of our knowledge, the highest for logistic regression models in the published literature for MIMIC cardiac patients (please refer to the discussion and eTable 2 in the supplement for the comparison).

*Table 1 Accuracy and AUC values for different outcomes and models.*

|  | EMR | GPT | EMR + GPT |
|---|---|---|---|
| 1-Year Mortality | 82.9% | 86.6% | 88.6% |
| 90-Day Mortality | 84.3% | 86.4% | 88.4% |
| 90-Day Readmission | 66.6% | 71.1% | 72.0% |

To understand why the GPT answers are so beneficial, we first compare the patient predictions from different models and then examine GPT´s assessment of the risk of death in more detail. Looking at the predictions for 1-year mortality from the EMR model alongside those from the EMR+GPT model (please refer to Figure 2, plots the mortality predictions of the two models), we see that they broadly agree on the risk of most patients, with a correlation of 0.85. In general, the models agree on which patients are low risk and which patients are high risk. However, in the upper left corner of the figure, there are some patients rated as very high risk by the EMR+GPT model but as very low risk by the EMR model. A manual review of discharge notes for the patients with the largest increase in predicted risk when the GPT answers are used in the model, reveals discussions about halting care, treating symptoms, providing palliative care, or refusing invasive treatment. Conversely, the patients showing the greatest decrease in predicted risk when discharge note information is included tend to have had major procedures that have gone well, like resection of metastatic cancer. We therefore reason that the EMR model "sees" that the patient is complex and has severe diagnoses, and thus it assigns the patient a high risk score. However, the EMR+GPT model, with access to the treatment and prognosis information in the discharge note, assigns a lower risk score for those patients where mitigating information is present in the discharge note.

To further understand the drivers of the predictions, we study GPTs answers to the question: "What is the patient's risk of death? (Rate no risk = 1 to very high risk = 100)". Interestingly, when GPT answers this question, it responds with one of only eight unique values (10, 20, 30, 50, 60, 70, 75, 80) for over 99% of patients. Figure 2 (right) shows the mortality percentage for patients receiving those eight risk scores. We see that as the risk score goes up, the mortality rate increases exponentially. A logistic regression trained on this one question alone has an AUC of 0.82 for 1-year mortality, which is competitive with the full EMR model.

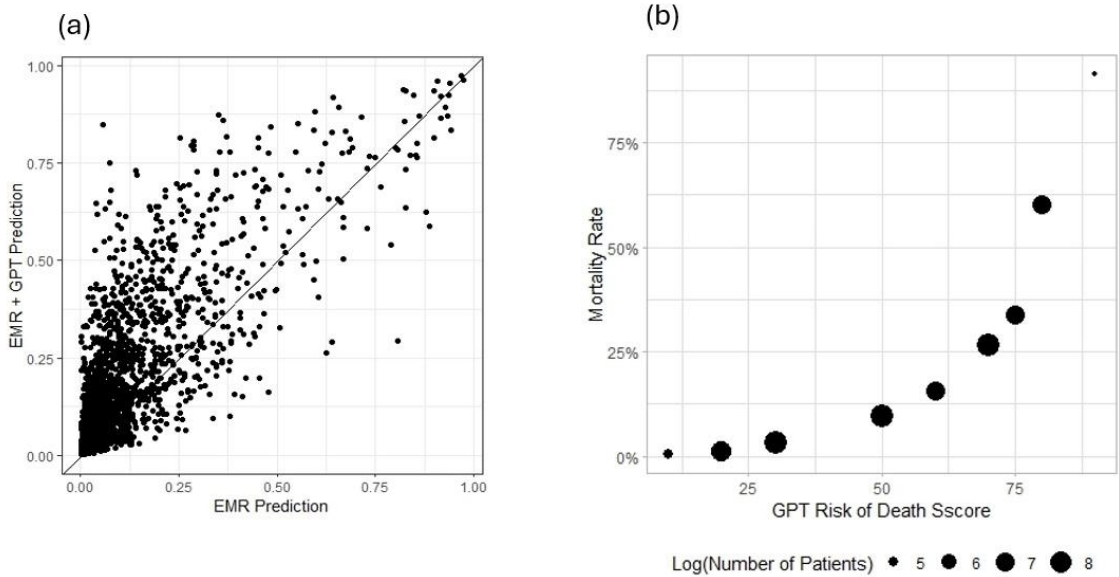

Figure 2. Comparison of EMR model predictions with EMR+GPT model predictions (left) and average mortality rate vs. GPT "risk of death" score (right)

We highlight the potential impact of the improved prediction accuracy by studying the positive predicted values of each model in predicting 90-day mortality and 1-year mortality. Table 2 summarizes the positive predictive value for each model and outcome across various cutoff thresholds. For example, if the goal is to identify patients with the highest mortality risk, of the top 2.5% highest risk patients based on the EMR+GPT model with 90-day and 1-year cutoffs, 59% die within 90 days and 77% die within one year. This compares to only 43% and 58% respectively, for the EMR-only model.

*Table 2* *Positive predictive value (hit rate) among the riskiest patients as identified for each model and outcome.*

| Outcome | Percentage of patients | EMR | GPT | EMR + GPT |
|---|---|---|---|---|
| 90-day Mortality | 2.50% | 43% | 59% | 63% |
|  | 5% | 31% | 45% | 49% |
|  | 10% | 26% | 32% | 32% |
|  | 20% | 18% | 20% | 21% |
| 1-year Mortality | 2.50% | 58% | 77% | 78% |
|  | 5% | 51% | 64% | 68% |
|  | 10% | 43% | 52% | 55% |
|  | 20% | 34% | 37% | 39% |
| 90-day Readmission | 2.50% | 21% | 17% | 17% |
|  | 5% | 17% | 18% | 16% |
|  | 10% | 13% | 16% | 18% |
|  | 20% | 11% | 16% | 16% |

**3. Discussion**

Our study focuses on the benefits of using LLMs, and GPT in particular, to synthesize unstructured data in medical records for patient outcome prediction. By building an expert driven prompt, we are able to leverage the unstructured data in care providers' discharge notes to significantly improve these predictions. This approach offers the potential to improve hospital operations and patient outcomes. Crucially, it is transparent, intuitive, and easy to implement.

Our study shows that GPTs answers to simple patient level questions are capable of stratifying patients according to their risk of mortality based on the text of the discharge summary. The information extracted from the LLM can also be used to improve traditional machine-learning models based on features from electronic medical records data. As hospitals improve their predictions of discharge patients' outcomes, they could then manage resources more effectively and schedule/prioritize follow-up care for the patients who need it most. This is particularly important in reducing avoidable readmission and improving patient care.

While AI has been demonstrated to improve predictive accuracy in healthcare[13,21,28], we argue that its true impact can only be achieved when it is implemented in clinical settings or used by care providers.

Our simple method is easy to automatically embed in an EMR or other medical information system, and it can easily deliver information to providers almost instantly and based solely on their medical note. We also acknowledge that algorithmic fairness considerations are critical to implementation in clinical settings. Our fairness analysis reveals that the benefit from using GPT's answers in the model features is shared across all demographic groups. Specifically, patients of both genders and patients both above and below 65 years of age show increases in model performance (MIMIC does not contain information on race or ethnicity). Further, the degree of benefit varies and is not uniformly larger for any one demographic group. For instance, women see greater performance gains when both GPT and tabular features are included in predicting 90-day mortality, whereas men benefit more in the 1-year mortality prediction. The details of these experiments are provided in eTable 3 in the supplement.

Finally, we note that our results are not sensitive to the population specification and can expand to other general purpose LLMs. For example, we reran the analyses on a broader sample of all patients who have any ICD-9 code with cardiac in the diagnosis title (please refer to eTable 4 in the supplement for a full list). The predictive performance for this more heterogeneous patient population were very similar. An important future research direction is to expand the analysis to a broader population, to further understand if the benefits of our approach extends to all hospitalizations. The EMR+GPT model is consistently two to four percentage points better than the EMR model at predicting mortality (as measured by the AUC), while the GPT-only model is equivalent in performance to the EMR model. Again, this shows that there is significant clinical information embedded in the discharge summary and that GPT is able to effectively extract and quantify the information across patient populations. We further note that in our expanded experiments we have used a number of different general-purpose LLMs, including many of GPT's competitors, and find similar performance, although GPT typically performs the best. This highlights that our approach of asking an LLM to read a discharge note and answer key questions is not limited to GPT but applies broadly to general purpose LLMs.

This method is a simple, transparent approach for incorporating generative AI into the medical prediction pipeline. It has the potential to perform as well as machine learning models based on tabular data while also being much simpler to implement and deploy. Alternatively, the information extracted from notes by the LLM can be used to augment existing models based on tabular EMR data. Physician notes are a rich source of qualitative data, and LLMs offer a method for distilling that information into an actionable quantitative prediction. This allows AI to complement human physicians by better utilizing the information-rich notes they write.

# Supplement for:

# Paging Dr. GPT: Extracting Information from Clinical Notes to Enhance Patient Predictions

1. **EMR Feature List**

eTable 1 summarizes the EMR data elements used in the study. Data elements with multiple observations over a patient's stay, such as vital signs and test results, are summarized and we use the minimum, maximum, and mean values as features.

*eTable 1* Summary of EMR features [An Excel Version of the Table is included with the submission]

| Group | Name | Description |
|---|---|---|
| Demographics | Gender | Gender of patient. |
| | Age at Discharge | Age of patient at discharge. |
| Comorbidities | Charlson Comorbidity Index | Weighted index to predict risk of death within 1 year of hospitalization for patients with specific comorbid conditions. Scores greater than 5 are |
| | Charlson Age Score | Number of decades after age 40 capped at 4. |
| | AIDS | Binary indicator of AIDS. |
| | Cerebrovascular Disease | Binary indicator of cerebrovascular disease. |
| | Chronic Pulmonary Disease | Binary indicator of chronic pulmonary disease. |
| | Congestive Heart Failure | Binary indicator of congestive heart failure. |
| | Dementia | Binary indicator of dementia. |
| | Diabetes without CC | Binary indicator of diabetes with carbohydrate counting. |
| | Diabetes with CC | Binary indicator of diabetes without carbohydrate counting. |
| | Malignant Cancer | Binary indicator of malignant cancer. |
| | Metastatic Solid Tumor | Binary indicator of metastatic solid tumor. |
| | Mild Liver Disease | Binary indicator of mild liver disease. |
| | Myocardial Infarct | Binary indicator of myocardial infarct. |
| | Paraplegia | Binary indicator of paraplegia. |
| | Peptic Ulcer Disease | Binary indicator of peptic ulcer disease. |
| | Peripheral Vascular Disease | Binary peripheral vascular disease. |
| | Renal Disease | Binary indicator of renal disease. |
| | Rheumatic Disease | Binary indicator of rheumatic disease. |
| | Severe Liver Disease | Binary indicator of severe liver disease. |
| Lab Tests | Anion Gap | electrolytes in blood. |
| | Lactage | Amount of lactic acid in blood. |
| | Urea Nitrogen | Amount of urea nitrogen in blood. |
| | Blood pH | pH measure of blood. |
| | White Blood Cell Count | Number of white blood cells in blood. |
| | Bicarbonate | Number of carbon dioxide in blood. |
| | Free Calcium | Amount of free calcium in blood (calcium not attached to proteins). |
| | Chloride | Amount of chloride in blood. |
| | Sodium | Amount of sodium in blood. |
| | Potassium | Amount of potassium in blood. |
| | Glucose | Amount of glucose in blood. |
| | PT | Prothrombin time (how long it takes blood to clot). |
| | PTT | Partial thrombin time. |
| | Alkaline Phosphatase | Amount of ALP in blood. |
| | Hematocrit | Percentage of red blood cells in blood. |
| | Albumin | Amount of albumin in blood. \midrule |

*eTable 1 cont.* Summary of EMR features

| Group | Name | Description |
|---|---|---|
| SAPSII* | SAPSII Score | Comorbidity score using the following parameters. |
| | PaO2/FiO2 Score | P/F ratio score; varying increments added for values above or below normal ratio of arterial oxygen partial pressure to fractional inspired |
| | SAPSII Age Score | Score from 0 to 18; varying increments added from ages 40 to 81+. |
| | Heart Rate Score | Varying increments added for rate above or below average heart rate |
| | SAPSII Probability | SAPSII probability |
| | Systolic Blood Pressure Score | Systolic blood pressure score; varying increments added for amounts greater or less than normal. |
| | Temperature Score | Temperature score; varying increments added for degrees greater or less than normal. |
| SOFA* | Glasgow coma score | Score that quantifies the extent of impaired consciousness with higher scores being more conscious. |
| | PaO2/FiO2 Ratio, No Vent | P/F ratio with no ventilation. |
| | PaO2/FiO2 Ratio, Vent | P/F ratio with ventilation. |
| | Dobutamine Rate | Rate of dobutamine administered. |
| | Dopamine Rate | Rate of dopamine administered. |
| | Epinephrine Rate | Rate of epinephrine administered. |
| | Norepinephrine Rate | Rate of norepinephrine administered. |
| | 24 Hour Urine Output | Urine output over 24 hours. |

## 2. Predictive Performance Comparison

We note that it is highly problematic to compare predictive performance across studies that do not use the exact same data (or at least use very similar populations), as any performance metric is highly dataset dependent. Therefore, as an indication only, we compare other studies that have used the MIMIC data and logistic regression to predict out-of-hospital mortality. One key consideration when predicting post-discharge mortality, is the inclusion or exclusion of patients not discharged alive. Including patients not discharged alive, in our experiments, artificially improves performance metrics (it is very easy to predict their mortality as they do not leave the hospital alive). It is unclear whether this criteria was applied in the comparison papers. Another key consideration is the patient cohort. A critically ill cohort is easier to predict that patients with more complex prognosis.

Only a handful of studies have used the MIMIC datasets to study out-of-hospital mortality for cardiac patients using logistic regression. eTable 2 summarizes these studies, as well as others that use the MIMIC data to predict out-of-hospital mortality using logistic regression for other cohorts. We note that our reported AUC is 0.874 for 90-mortality and 0.871 for 1-year mortality, higher than those reported by [1-8]. We however note that papers [2-4] additionally apply more advanced machine learning algorithm such as boosting and random forest, and report higher AUC for those methods compared to logistic regression.

*eTable 2 AUC values of studies using logistic regression to predict out-of-hospital mortality*

| Authors | Year | Population | Mortality target | AUC |
|---|---|---|---|---|
| Lin et al. [1] | 2024 | Myocardial Infarction | 30-day | 0.84 |
| | | | 30-day | 0.82 |
| | | | 180-day | 0.75 |
| | | | 1-year | 0.70 |
| Li et al. [2] | 2023 | Heart Failure | > 1-year | 0.72 |
| Shi et al. [3] | 2024 | Heart Failure with Clostridioides difficile infection | 28-day | 0.76 |
| Shin et al. [4] | 2024 | Heart Failure with Preserved Ejection Fraction | 30-day | 0.83 |
| | | | 1-year | 0.75 |
| Yag et al. [5] | 2023 | Trauma | 90-day | 0.81 |
| Zhang et al. [6] | 2024 | Sepsis | 28-dy | 0.83 |
| Hou et al. [7] | 2020 | Sepsis | 30-day | 0.82 |
| Su et al. [8] | 2022 | Sepsis | 30-day | 0.75 |

### 3. AUC Values by Gender and Age Bucket

We first calculate the AUC value by gender and two age buckets, below 65 years old and at or above 65 years old, and then calculate the performance difference by gender and age bucket. eTable 3 summarizes these results.

We first note, that across outcomes, there are small performance differences both by gender and by age bucket, when we use only the tabular EMR data. When we only use GPT's answer, in four cases out of six the differences increase, in two cases they decrease. When we combine the tabular data with GPT's answers and compare the age and gender differences with using EMR data only, half the time the differences are bigger, half the time the differences are smaller. We however note that in all cases, going from only using EMR tabular data to using both EMR and GPT data improves the AUC.

**eTable 3**. AUC broken down by gender (male and female) and age (65 and above, and younger than 65).

| | AUC Values by Gender | | | | | |
|---|---|---|---|---|---|---|
| | 1-year mortality | | 90-day mortality | | 90-day readmission | |
| | Female | Male | Female | Male | Female | Male |
| EMR | 82.3% | 82.6% | 82.4% | 85.0% | 63.8% | 67.7% |
| GPT | 84.5% | 87.6% | 84.0% | 87.4% | 68.3% | 72.5% |
| EMR + GPT | 87.1% | 89.1% | 86.7% | 89.0% | 69.1% | 73.4% |
| | Absolute Differences in AUC Values by Gender | | | | | |
| | 1-year mortality | | 90-day mortality | | 90-day readmission | |
| | Female - Male | | Female - Male | | Female - Male | |
| EMR | -0.3% | | -2.6% | | -3.9% | |
| GPT | -3.1% | | -3.4% | | -4.2% | |
| EMR + GPT | -2.0% | | -2.3% | | -4.3% | |
| | AUC Values by Age Bucket | | | | | |
| | 1-year mortality | | 90-day mortality | | 90-day readmission | |
| | Under 65 | 65+ | Under 65 | 65+ | Under 65 | 65+ |
| EMR | 81.7% | 81.9% | 84.9% | 82.9% | 68.7% | 65.9% |
| GPT | 87.7% | 86.0% | 87.4% | 85.5% | 72.2% | 70.8% |
| EMR + GPT | 88.7% | 87.7% | 88.0% | 86.9% | 72.7% | 71.9% |
| | Absolute Differences between AUC Values by Age Bucket | | | | | |
| | 1-year mortality | | 90-day mortality | | 90-day readmission | |
| | Under 65 - 65+ | | Under 65 - 65+ | | Under 65 - 65+ | |
| EMR | -0.2% | | 2.0% | | 2.8% | |
| GPT | 1.7% | | 1.9% | | 1.4% | |
| EMR + GPT | 1.0% | | 1.1% | | 0.8% | |

4. **ICD Codes**

eTable 4 summarizes the ICD-9 and ICD-10 codes used to identify the broader cardiac cohort.

*eTable 4* List of ICD codes used to identify a broader cardiac cohort [An Excel Version of the Table is included with the submission]

| ICD Code | Long Title | ICD Code | Long Title | ICD Code | Long Title |
| --- | --- | --- | --- | --- | --- |
| 4280 | Congestive heart failure unspecified | I2510 | Atherosclerotic heart disease of native coronary artery without angina pectoris | 412 | Old myocardial infarction |
| 42789 | Other specified cardiac dysrhythmias | 4168 | Other chronic pulmonary heart diseases | I110 | Hypertensive heart disease with heart failure |
| I252 | Old myocardial infarction | I214 | Non-ST elevation (NSTEMI) myocardial infarction | 9971 | Cardiac complications, not elsewhere classified |
| I130 | Hypertensive heart and chronic kidney disease with heart failure and stage 1 through stage 4 chronic kidney disease | Z8249 | Family history of ischemic heart disease and other diseases of the circulatory system | 42833 | Acute on chronic diastolic heart failure |
| 42832 | Chronic diastolic heart failure | 42823 | Acute on chronic systolic heart failure | 4254 | Other primary cardiomyopathies |
| I5033 | Acute on chronic diastolic (congestive) heart failure | 42822 | Chronic systolic heart failure | 4275 | Cardiac arrest |
| V4501 | Cardiac pacemaker in situ | V173 | Family history of ischemic heart disease | I5023 | Acute on chronic systolic (congestive) heart failure |
| 78551 | Cardiogenic shock | R570 | Cardiogenic shock | I5032 | Chronic diastolic (congestive) heart failure |
| I5022 | Chronic systolic (congestive) heart failure | 42821 | Acute systolic heart failure | 4148 | Other specified forms of chronic ischemic heart disease |
| I255 | Ischemic cardiomyopathy | I21A1 | Myocardial infarction type 2 | I5021 | Acute systolic (congestive) heart failure |
| I509 | Heart failure, unspecified | 41189 | Other acute and subacute forms of ischemic heart disease, other | V4502 | Automatic implantable cardiac defibrillator in situ |
| I25110 | Atherosclerotic heart disease of native coronary artery with unstable angina pectoris | Z950 | Presence of cardiac pacemaker | E8790 | Cardiac catheterization as the cause of abnormal reaction of patient, or of later complication |
| 42831 | Acute diastolic heart failure | I429 | Cardiomyopathy, unspecified | V433 | Heart valve replaced by other means |
| 42843 | Acute on chronic combined systolic and diastolic heart failure | I248 | Other forms of acute ischemic heart disease | 4233 | Cardiac tamponade |
| I469 | Cardiac arrest, cause unspecified | 41041 | Acute myocardial infarction of other inferior wall, initial episode of care | I5031 | Acute diastolic (congestive) heart failure |
| I314 | Cardiac tamponade | I25118 | Atherosclerotic heart disease of native coronary artery with other forms of angina pectoris | 99672 | Other complications due to other cardiac device |
| I132 | Hypertensive heart and chronic kidney disease with heart failure and with stage 5 chronic kidney disease, or end stage renal disease | 41011 | Acute myocardial infarction of other anterior wall, initial episode of care | Z95810 | Presence of automatic (implantable) cardiac defibrillator |

*eTable 4 cont* List of ICD codes used to identify a broader cardiac cohort

| ICD Code | Long Title | ICD Code | Long Title | ICD Code | Long Title |
|---|---|---|---|---|---|
| V422 | Heart valve replaced by transplant | Z8674 | Personal history of sudden cardiac arrest | I97190 | Other postprocedural cardiac functional disturbances following cardiac surgery |
| Z952 | Presence of prosthetic heart valve | I2119 | STEMI myocardial infarction involving other coronary artery of inferior wall | V1749 | Family history of other cardiovascular diseases |
| Y840 | Cardiac catheterization as the cause of abnormal reaction of the patient | Z45018 | Encounter for adjustment and management of other part of cardiac pacemaker | 42830 | Diastolic heart failure, unspecified |
| I428 | Other cardiomyopathies | I5043 | Acute on chronic combined systolic and diastolic heart failure | V1253 | Personal history of sudden cardiac arrest |
| I420 | Dilated cardiomyopathy | 79431 | Nonspecific abnormal ECG [EKG] | I5020 | Unspecified systolic (congestive) heart failure |
| I2109 | STEMI myocardial infarction involving other coronary artery of anterior wall | 41091 | Acute myocardial infarction of unspecified site, initial episode of care | I5030 | Unspecified diastolic (congestive) heart failure |
| T8111XA | Postprocedural cardiogenic shock, initial encounter | 42820 | Systolic heart failure, unspecified | I462 | Cardiac arrest due to underlying cardiac condition |
| 42842 | Chronic combined systolic and diastolic heart failure | 4299 | Heart disease, unspecified | Z4502 | Encounter for adjustment and management of automatic implantable cardiac defibrillator |
| 42989 | Other ill-defined heart diseases | E9429 | Other agents primarily affecting the cardiovascular system causing adverse effects in therapeutic use | V5331 | Fitting and adjustment of cardiac pacemaker |
| I468 | Cardiac arrest due to other underlying condition | Z953 | Presence of xenogenic heart valve | 99661 | Infection and inflammatory reaction due to cardiac device, implant, and graft |
| E9420 | Cardiac rhythm regulators causing adverse effects in therapeutic use | I513 | Intracardiac thrombosis, not elsewhere classified | 4293 | Cardiomegaly |
| 41001 | Acute myocardial infarction of anterolateral wall, initial episode of care | 99801 | Postoperative shock, cardiogenic | 4359 | Unspecified transient cerebral ischemia |
| I213 | STEMI myocardial infarction of unspecified site | 42841 | Acute combined systolic and diastolic heart failure | R9431 | Abnormal electrocardiogram [ECG] [EKG] |
| 40491 | Hypertensive heart and chronic kidney disease, unspecified | N280 | Ischemia and infarction of kidney | 41021 | Acute myocardial infarction of inferolateral wall, initial episode of care |
| V5332 | Fitting and adjustment of automatic implantable cardiac defibrillator | 41031 | Acute myocardial infarction of inferoposterior wall, initial episode of care | I498 | Other specified cardiac arrhythmias |

*eTable 4 cont* List of ICD codes used to identify a broader cardiac cohort

| ICD Code | Long Title | ICD Code | Long Title | ICD Code | Long Title |
|---|---|---|---|---|---|
| I97191 | Other postprocedural cardiac functional disturbances following other surgery | T827XXA | Infection and inflammatory reaction due to other cardiac and vascular devices, implants, initial encounter | I421 | Obstructive hypertrophic cardiomyopathy |
| 99602 | Mechanical complication due to heart valve prosthesis | 39891 | Rheumatic heart failure (congestive) | I5041 | Acute combined systolic and diastolic heart failure |
| 42511 | Hypertrophic obstructive cardiomyopathy | I5042 | Chronic combined systolic and diastolic heart failure | 99601 | Mechanical complication due to cardiac pacemaker (electrode) |
| T82857A | Stenosis of other cardiac prosthetic devices, implants and grafts, initial encounter | I517 | Cardiomegaly | 4255 | Alcoholic cardiomyopathy |
| I2722 | Pulmonary hypertension due to left heart disease | V151 | Personal history of surgery to heart and great vessels, presenting hazards to health | 40291 | Unspecified hypertensive heart disease with heart failure |
| I97820 | Postprocedural cerebrovascular infarction following cardiac surgery | I119 | Hypertensive heart disease without heart failure | 4169 | Chronic pulmonary heart disease, unspecified |
| Z954 | Presence of other heart-valve replacement | I422 | Other hypertrophic cardiomyopathy | T796XXA | Traumatic ischemia of muscle, initial encounter |
| 99671 | Other complications due to heart valve prosthesis | I2111 | STEMI myocardial infarction involving right coronary artery | 99604 | Mechanical complication of automatic implantable cardiac defibrillator |
| Y718 | Miscellaneous cardiovascular devices associated with adverse incidents | 4251 | Hypertrophic obstructive cardiomyopathy | I97611 | Postprocedural hemorrhage of a circulatory system organ or structure following cardiac bypass |
| I253 | Aneurysm of heart | Y713 | Surgical instruments, materials and cardiovascular devices associated with adverse incidents | I259 | Chronic ischemic heart disease, unspecified |
| 4289 | Heart failure, unspecified | Y712 | Prosthetic and other implants, materials and accessory cardiovascular devices associated with adverse incidents | 39890 | Rheumatic heart disease, unspecified |
| 4258 | Cardiomyopathy in other diseases classified elsewhere | V1741 | Family history of sudden cardiac death (SCD) | I2129 | STEMI myocardial infarction involving other sites |
| I50810 | Right heart failure, unspecified | 41410 | Aneurysm of heart (wall) | 4259 | Secondary cardiomyopathy, unspecified |
| E8706 | Accidental cut, puncture, perforation during heart catheterization | 2127 | Benign neoplasm of heart | T82897A | Other specified complication of cardiac prosthetic devices, implants and grafts, initial encounter |

continued...

*eTable 4 cont* List of ICD codes used to identify a broader cardiac cohort

| ICD Code | Long Title | ICD Code | Long Title | ICD Code | Long Title |
|---|---|---|---|---|---|
| R011 | Cardiac murmur, unspecified | 4257 | Nutritional and metabolic cardiomyopathy | I97710 | Intraoperative cardiac arrest during cardiac surgery |
| Y711 | Therapeutic (nonsurgical) cardiovascular devices associated with adverse incidents | 4358 | Other specified transient cerebral ischemias | I43 | Cardiomyopathy in diseases classified elsewhere |
| E8583 | Accidental poisoning by agents primarily affecting cardiovascular system | 7852 | Undiagnosed cardiac murmurs | I97790 | Other intraoperative cardiac functional disturbances |
| I499 | Cardiac arrhythmia, unspecified | Z45010 | Encounter for checking and testing of cardiac pacemaker pulse generator [battery] | I2102 | STEMI myocardial infarction involving left anterior descending coronary artery |
| 42518 | Other hypertrophic cardiomyopathy | E9421 | Cardiotonic glycosides and drugs of similar action causing adverse effects in therapeutic use | T826XXA | Infection and inflammatory reaction due to cardiac valve prosthesis |
| K55019 | Acute (reversible) ischemia of small intestine, extent unspecified | 41051 | Acute myocardial infarction of other lateral wall, initial episode of care | I426 | Alcoholic cardiomyopathy |
| I5189 | Other ill-defined heart diseases | T82867A | Thrombosis due to cardiac prosthetic devices, implants and grafts, initial encounter | 41092 | Acute myocardial infarction of unspecified site, subsequent episode of care |
| D151 | Benign neoplasm of heart | | | | |